# Comparative study of microgrid optimal scheduling under multi-optimization algorithm fusion

Hongyi Duan [1,a]
[1]Xi'an Jiaotong University
Faculty of Electronic and Information Engineering
Xi'an, China
[a]dann_hiroaki@ieee.org

Qingyang Li [1,b]
[1]Xi'an Jiaotong University
Faculty of Electronic and Information Engineering
Xi'an, China
[b]likon2101@ieee.org

Yuchen Li [1,c]
[1]Xi'an Jiaotong University
Faculty of Electronic and Information Engineering
Xi'an, China
[c]yuchenli@ieee.org

Tianjiao Ji [1,d]
[1]Xi'an Jiaotong University
Faculty of Electronic and Information Engineering
Xi'an, China
[d]SherlockJi@ieee.org

Jianan Zhang [2,e*]
[2]Shanghai University of Finance and Economics University
School of Mathematica
Shanghai, China
Corresponding author: [e*]zjaqifei@ieee.org

Yuming Xie [3,f*]
[3]National University of Defense Technology
Faculty of Electronic Science
Changsha, China
xieyuming@ieee.org

Hongyi Duan, Qingyang Li and Yuchen Li Contributed equally to this article and should be considered co-first authors

*Abstract*—As global attention on renewable and clean energy grows, the research and implementation of microgrids become paramount. This paper delves into the methodology of exploring the relationship between the operational and environmental costs of microgrids through multi-objective optimization models. By integrating various optimization algorithms like Genetic Algorithm, Simulated Annealing, Ant Colony Optimization, and Particle Swarm Optimization, we propose an integrated approach for microgrid optimization. Simulation results depict that these algorithms provide different dispatch results under economic and environmental dispatch, revealing distinct roles of diesel generators and micro gas turbines in microgrids. Overall, this study offers in-depth insights and practical guidance for microgrid design and operation.

*Keywords*—Microgrid, multi-target, Optimization Algorithm, Multi-objective Dispatchs

## I. INTRODUCTION

### A. Introduction to Microgrids

With the transformation of the global energy structure and the increased reliance on renewable energy sources, microgrids (microgrids) are gradually becoming an important part of the power system. Microgrids are autonomous, controllable, and flexible, and can optimize energy use, increase the stability of energy supply, and reduce environmental impacts[1]. At the level of optimizing energy use, microgrids can achieve local generation and consumption of energy and increase the efficiency of energy use[2]. At the level of guaranteeing power supply reliability, microgrids can supply power independently when the main grid fails, ensuring the stable operation of critical loads[3]. At the level of promoting access to renewable energy, microgrids provide a platform that makes it easier for renewable energy sources such as solar and wind to access the grid[4].

The operation mode of microgrids is categorized into grid-connected mode and islanded mode, where the islanded mode refers to the independent operation of microgrids from the main grid. In this mode, the microgrid needs to ensure the internal supply and demand balance, and the factors to be considered include: the reliability of the power supply, the change of the load, the state of the energy storage equipment, and the management and control strategy within the microgrid[5].

The research field of microgrids is wide-ranging, covering a variety of aspects from system design, control strategies to economic analysis. In recent years, researchers have conducted in-depth studies on the stability, reliability, and economy of microgrids[6]. In addition, with the development of digital and communication technologies, the management and control strategies of microgrids have been revolutionized[7]. Many studies have also focused on how to better integrate renewable energy sources, how to design more efficient energy storage systems, and how to optimize the operation and dispatch of microgrids[8].

*B. Research method and purpose of this paper*

Based on the power characteristics of micro power supply, this paper discusses the dynamic economic load distribution problem of microgrid in islanding operation mode, mainly focuses on three indexes of microgrid's economic efficiency, environmental friendliness and stability, and at the same time tries to incorporate many other factors such as outage loss, access capacity as auxiliary considerations, and thus establishes a multi-objective optimal scheduling model of microgrid. In addition, this paper adopts multi-objective genetic algorithm, multi-objective particle swarm algorithm, multi-objective ant colony algorithm, and multi-objective simulated annealing algorithm to solve the model, in order to filter out the optimization algorithm that can take into account of the convergence speed and exploration ability.

II. OVERVIEW OF THE ALGORITHMS, SYSTEMS, AND METHODS USED IN THIS PAPER

*A. Optimization algorithms and their multi-objective improvement*

*1) Genetic Algorithm and Multi-objective Genetic Algorithm*

Genetic Algorithm (GA) is a search algorithm that simulates natural selection and genetic mechanism, which is suitable for finding solutions to optimization problems in a large search space. The basic steps are as follows:

①Initialization: randomly generate a population $P$ of size $N$ such that each of $x_i$ in $P(t) = \{x_1, x_2, ..., x_N\}$ is a candidate solution

②Evaluate fitness: use the fitness function $f$ to evaluate the fitness for each candidate solution, i.e. $f(x_i)$

③Selection: select $N$ individuals into the next generation based on the fitness, that is $P(t) = \{x_{i1}, x_{i2}, ..., x_{iN}\}$, using roulette selection.

④Crossover: randomly select two parent individuals and crossover based on crossover probability to generate offspring.

⑤Mutation: make small changes to individuals based on the mutation probability.

⑥Termination: when a preset number of iterations or other stopping criterion is reached.

Multi-objective genetic algorithms aim to optimize multiple objective functions with the goal of usually finding a solution on the Pareto frontier. Given $m$ objective functions $f_1, f_2, ... f_m$, and we define a solution $x$ that outperforms $y$, which means if on all objectives, $x$ does not outperform $y$, or on at least one objective, $x$ outperforms $y$. The basic steps are highly similar to those of genetic algorithms, with the main differences being that the selection process uses a non-dominated sorting method to rank the populations and the need to update the current Pareto frontier solution at each generation at the end of the process. In addition, in order to maintain the diversity of the population, MOGA usually uses diversity strategies such as crowding distance.

*2) Particle swarm optimization algorithm and multi-objective particle swarm optimization algorithm*

Particle swarm optimization simulates the behavior of a flock of birds foraging for food, in this model each particle represents a point in the solution space. The basic steps are as follows

①Initialize n particles. Each particle i has an initial position $X_i(0) = rand()$ and an initial velocity $V_i(0) = rand()$ in the solution space.

②In each iteration, update the particle velocity following

$$V_i(t+1) = \omega V_i(t) + P_{diff} rand() + G_{diff} rand()$$
$$P_{diff} = c_1(P_{best,i} - X_i(t)) \quad (1)$$
$$G_{diff} = c_2(G_{best,i} - X_i(t))$$

formula in order to update the particle velocity following the formula $X_i(t+1) = X_i(t) + V_i(t+1)$, which indicates the default interval between each update is one unit of time.

③The algorithm continues to iterate until a predetermined number of iterations or other termination criteria are reached.

Where the meaning of the parameters are: $G_{best}$ for the global best, $P_{best,i}$ for the best historical position of the particle, $\omega$ for the weight, $c_1, c_2$ for the learning factor

The core idea of the multi-objective particle swarm optimization algorithm likewise turns out to be finding the Pareto frontier of the multi-objective problem. The steps are similar to those of the particle swarm optimization algorithm, with the difference that before iteration the algorithm sorts the particles based on non-dominated relationships, and the particle update formula $G_{best}$ becomes a non-dominated solution selected from an external archive (the algorithm uses an external archive in order to preserve the historical non-dominated solutions).

*3) Ant Colony Algorithm and Multi-Objective Ant Colony Algorithm*

The Ant Colony Algorithm is a heuristic optimization technique inspired by the behavior of natural ant colonies in finding food paths. The algorithm is based on ants communicating using pheromones and working together to find a food source. The algorithm flow can be as follows.

①Initialization: let the ant colony size be m and the initialized pheromone concentration be $\tau_{ij}(0)$.

②Construct the solution: the next node is selected by the ants based on the pheromone concentration $\tau_{ij}$ and the heuristic function $\eta_{ij}(0)$ (e.g., the reciprocal of the distance). The selection probability formula is:

$$P_{ij}(t) = \frac{\tau_{ij}^{\alpha} \eta_{ij}^{\beta}}{\sum_{k \in allowed} \tau_{ik}^{\alpha} \eta_{ik}^{\beta}} \quad (2)$$

where $\alpha, \beta$ are the regulation parameters of pheromone importance and heuristic function importance respectively

③Update pheromone: the pheromone will be volatilized over time, and at the same time updated according to the length of the path found by the ants, with the formula

$$\tau_{ij}(t+1) = (1-\rho)\tau_{ij}(t) + \Delta\tau_{ij}(t) \quad (3)$$

where $\rho$ is the volatilization coefficient, which $\Delta\tau_{ij}(t)$ is the total pheromone left by the ants on the $(i,j)$ edge in the $t$ iteration

④The algorithm terminates after reaching the maximum number of iterations

The multi-objective version of the ACO algorithm aims to find the set of Pareto optimal solutions to deal with possible conflicts between multiple objectives. In contrast to traditional ACO algorithms, it is desirable to incorporate a solution construction and evaluation mechanism before pheromone follow-up, and pheromone updating in multi-objective scenarios may be based on non-dominated ranking or other mechanisms to identify high-quality non-dominated solutions, in addition to possibly comparing and updating the set of non-dominated solutions with the current set of non-dominated solutions for each new solution found.

*4) Simulated Annealing Algorithm and Multi-Objective Simulated Annealing Algorithm*

The simulated annealing algorithm is inspired by the physical annealing process in which the material is heated at high temperatures and then slowly cooled to reach a low energy state. The brief steps are as follows:

①Select initial solution $s$ and initial temperature $T$.

②Iterative process: Multiple iterations are performed at the current temperature, with each iteration attempting a new solution $s'$.

③Acceptance criterion: If the energy (or cost) of the new solution is low, accept it. Otherwise, accept the worse solution with the probability described in this equation

$$P(e,e',T) = e^{(\frac{e'-e}{T})} \quad (4)$$

where $e, e'$ are the energies of the current solution and the new solution, respectively.

④Reduce temperature: reduce the temperature as a decreasing function of $T = \alpha T, \alpha \in (0,1)$.

⑤ Terminate: terminate the algorithm if the temperature falls below a predetermined threshold or if the solution remains unchanged for a long time.

The multi-objective simulated annealing algorithm is designed to deal with problems with multiple optimization objectives and find the Pareto optimal set of solutions. Its differences from the standard simulated annealing algorithm can be found in the previous section on the differences between the three algorithms and their multi-objective improvement algorithms.

*B. Overview of renewable energy based microgrid power systems*

*1) General Architecture*

In this paper, a microgrid architecture is used as shown in Fig. 1, and all the contained units are shown in the figure. By default the system is islanded and the internal loads of the microgrid are powered by FC.

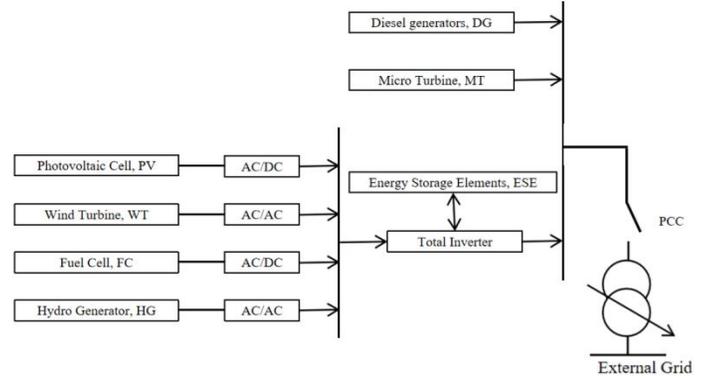

Figure 1. Microgrid Overall Structure

*2) Clean energy related modules*

The power generation efficacy of PV cells is directly related to the intensity of solar irradiation received and they are mostly operated in maximum power point tracking (MPPT) mode to optimize the capacity. Referring to the model optimized in the literature[9], we take the following output power equation in Table 1

Table 1. Photovoltaic Cell Related Expressions and Parameters

| Output power formula | $P_{real} = \zeta \eta_m A_p \eta_p \cos\theta$ (5) | |
|---|---|---|
| Parameter Meaning | $P_{real}$ | Actual output power |
| | $\zeta$ | light intensity |
| | $\eta_m$ | Efficiency in maximum power point tracking mode |
| | $A_p$ | Panel area |
| | $\theta$ | sun exposure angle |

The efficiency and power of hydroelectricity is related to the friction coefficient, flow rate, and height. Table 2 below shows some key equations describing the modeling of hydroelectric power generation.

Table 2. Hydropower Related Expressions and Parameters

| Relevant | Power formula | $P = \rho g Q H \eta_h$ (6) |

| formulas | Turbine efficiency formula | $\eta_t = \dfrac{\rho g Q H - P_{mec}}{\rho g Q H}$ (7) |
|---|---|---|
| | turbomachinery loss | $P_{mec} = f_{t-w} v_w^2 + f_{t-w} \omega_{sha}^2$ (8) |
| Meaning of key parameters | $P_{mec}$ | Power lost by turbomachinery |
| | $Q$ | water flow |
| | $f_{t-w}$ | Friction coefficients between turbine and water and turbine and shaft |
| | $v_w^2, \omega_{sha}^2$ | Water speed, turbine speed |

The power of the wind turbine with is mainly determined by the wind speed, and its power output model is described in Table 3.

Table 3. Wind Turbine Related Expressions and Parameters

| Power output formula | $P_{WT} = \begin{cases} 0 & v < v_{ci},\ v > v_{co} \\ \dfrac{v^3 - v_{ci}^3}{v_r^3 - v_{ci}^3} P_r & v_{ci} \le v \le v_r \\ P_r & v_r \le v \le v_{co} \end{cases}$ (9) | |
|---|---|---|
| Parameter Meaning | $P_{WT}, P_r$ | Actual power, rated power |
| | $v_{ci}, v_{co}, v_r$ | Cut-in wind speed, cut-out wind speed, rated wind speed are taken in the text as: 3 m/s、25 m/s、15 m/s |

### 3) Conventional energy-related modules

The micro gas turbine is added to this system as a controllable module, and based on the literature[10], whose fuel cost is related to the working efficiency, we can get the improved equation as described in Table 4

Table 4. Microturbine Related Expressions and Parameters

| Relevant formulas | Fuel cost formula | $F = \dfrac{CP}{LHV\eta}$ (10) |
|---|---|---|
| | Expression of working efficiency in relation to output power | $\eta = \sum_{i=1}^{4}\left[\dfrac{0.372}{1+\left(\dfrac{i}{3.445}\right)^{3.529}}\right] * \left(\dfrac{PM}{65}\right)^i$ (11) |
| Meaning of selected parameters | $C$ | Natural gas prices, in this paper, are taken to be $0.31USD/m^3$ |
| | $LHV$ | The calorific value of natural gas, in this paper, is taken as $1000 Btu/ft^3$ |

The cost of a diesel generator can be described by its consumption characteristic function, and its fuel cost can be fitted by the following function in Table 5, which is taken as $n = 4$ in this paper。

Table 5. Diesel Generator Related Expressions and Parameters

| Consumption characteristic function | $F = \sum_{i=1}^{n} P^i * [0.284 - 1.426 \lg(i)]$ (12) | |
|---|---|---|
| Parameter Meaning | $F, P$ | Fuel Cost, Output Power |

### 4) Battery Modules

Given the high instability of clean energy sources, a battery module is introduced in this system with the intention of buffering the presence of uncertain outputs to improve the reliability of the microgrid. We use the following model to describe the charging and discharging of the battery.

Table 6. Battery Storage Related Expressions and Parameters

| Charge and Discharge Expressions | Charging electric | $\begin{cases} E(t) = E(t-1) + \Delta E \\ \Delta E = \left[P_{total}(t) - \dfrac{P_{load}(t)}{\eta_{inv}}\right]\eta_{sb}\Delta t \end{cases}$ (13) |
|---|---|---|
| | Electrical discharge | $\begin{cases} E(t) = E(t-1) - \Delta E \\ \Delta E = \left[\dfrac{P_{load}(t)}{\eta_{inv}} - P_{total}(t)\right]\eta_{sb}\Delta t \end{cases}$ (14) |
| Meaning of selected parameters | $E(t)$ | Capacity of the battery at a given moment |
| | $P_{total}(t)$ $P_{load}(t)$ | The total output load of the power supply, and total load at a certain point in time |
| | $\eta_{inv}, \eta_{sb}$ | Inverter efficiency, charge/discharge efficiency |

### C. General Description of Multi-objective Optimization Problems

A multi-objective optimization problem can be defined as: given a decision vector $\vec{X} = \{x_1, x_2, ..., x_n\}$ and an objective function $f_i(x): R^n \to R, i = 1, 2, ..., m$ along with other constraints, thus seeking a solution set $\vec{X}^*$ such that each solution $\vec{x}^*$, satisfies all the constraints and achieves a better level of optimization for all the objective functions. This process can be simplified as:

$$\min or \max f(x) = \{f_1(x), f_2(x), ..., f_m(x)\}$$
$$s.t. \begin{cases} g_j(x) \le 0, j = 1, 2, ..., p \\ h_k(x) = 0, k = 1, 2, ..., q \end{cases} \quad (15)$$

where $f(x), g_j(x), h_k(x)$ are the objective function vector, the inequality constraint, and the equality constraint, respectively. $x$ must be in the solution space $X$.

### III. A MATHEMATICAL MODEL FOR OPTIMAL OPERATION OF MULTI-OBJECTIVE MICROGRIDS

### 1) General Architecture

#### A. Operating Costs

The operating cost of microgrid is mainly categorized into generation cost as well as interruption cost. So the minimum value of operating cost can be described as:

$$\min F = \sum_{t=1}^{T}\sum_{i}^{N} CO_{i,t}(P_{i,t}) + IR_t \quad (16)$$

Where $CO_{i,t}$ is the generation cost of the micropower $i$ source at the moment $t$, $IR_t$ is the interruptible cost of the microgrid $t$, and $P_{i,t}$ is the generation power of the micropower $i$ source at the moment $t$. However, in practice, many other factors as well as accidental factors need to be considered, so the following stochastic correction factors are considered to be introduced:

$$\varepsilon = (1+\alpha), \alpha \sim Uniform(0, 0.01) \quad (17)$$

Thus it follows that:

$$\min F = \left[\sum_{t=1}^{T}\sum_{i}^{N} CO_{i,t}(P_{i,t}) + IR_t\right]\varepsilon \quad (18)$$

*1) Power generation cost*

Since hydroelectricity, photovoltaic, and wind power are all clean energy sources, so this paper for the cost of power generation is mainly considered only the depreciation of equipment and maintenance costs, micro-power unit time depreciation costs can be expressed as:

$$IV_{i,t} = \frac{C_{INS,i}}{8760 P_{r,i} f_{c,i}} * \frac{d(1+d)^m}{(1+d)^m - 1} \quad (19)$$

where $C_{INS,i}, P_{r,i}, f_{c,i}$ is the landed cost of the $i$ th micropower supply, the power rating, and the capacity factor, respectively. $d$ is the depreciation factor, and $m$ is the service life. The maintenance cost is:

$$OM_{i,t}(t) = K_{m,i}(t) * P_{i,t}(t) \quad (20)$$

which $K_{m,i}(t), P_{i,t}(t)$ are the single maintenance cost of the first micro power supply and the probability of needing maintenance, respectively. All can be roughly considered as an increasing function of $t$, after experimentation, in this paper we take the following function in order to achieve the best results:

$$OM_{i,t}(t) = \left[\frac{3014}{3125}K_{m,i}(0)e^{\frac{286}{3200}t}\right] * \left[(1 + \frac{1}{502}t + \frac{1}{3398}t^2)P_{i,t}(0)\right] \quad (21)$$

Introducing the stochastic correction factor:

$$\varepsilon_i = (1+\alpha_i), \alpha_i \sim Uniform(0, 0.01), i = 1, 2 \quad (22)$$

in the same way yields the total generation cost as:

$$CO_{i,t} = IV_{i,t}\varepsilon_1 + OM_{i,t}\varepsilon_2 \quad (23)$$

*2) Interruptible costs*

In the face of power shortages during microgrid islanding operations, we propose a strategy that combines the urgency of loads with their historical interruption history. Loads are categorized as critical, important, and common with a score reflecting past interruption history. When a shortage of power supply is predicted, priority is given to interrupting ordinary loads with high historical scores, followed by other ordinary loads, then important loads, and finally critical loads.

From this, the economic compensation is given by Eq:

$$IR = A + BP_{IL}L + CH + D(P_{IL}H)^2 \quad (24)$$

$P_{IL}$ is the interrupt power, $L$ is a priority factor (critical=3, important=2, normal=1), and $H$ represents the historical interrupt score, ranging from 0 (no historical interrupts) to 1 (frequent interrupts)

*B. Environmental costs*

Referring to the study in the paper[11], We simply consider the emission purification cost of $CO_2, SO_2, NO_x$ as. Without considering the clean energy, the objective function for minimizing the environmental cost is

$$\min CE = \sum_{t=1}^{T}\sum_{j=1}^{K}\alpha_j \sum_{i=1}^{N}\beta_{ij}P_{i,t} \quad (25)$$

Where $K$ is the type of pollutant, is the unit cost of treatment for the gas, $\alpha_j$ taking into account the large fluctuations in the market also introduced a stochastic correction factor

$$\alpha_j = \alpha_j(0)\varepsilon_3, \varepsilon_3 = (1+\alpha_3), \alpha_3 \sim Uniform(0, 0.08) \quad (26)$$

In addition, the $\beta_{ij}$ emission factors for pollutants emitted at the time $P_i$ of output for different electricity production $j$ methods.

*C. Constraints*

For microgrid power balance first various constraints are described and the bundle matrix is defined to simplify the expression

$$\vec{X} = \begin{bmatrix} \vec{P} \\ \vec{BS} \\ \vec{\eta} \end{bmatrix} = \begin{bmatrix} P_i, P_{line} \\ P_{BS}, E_{BS} \\ \eta_+, \eta_- \end{bmatrix} \quad (27)$$

$\vec{P}$ contains the micropower output microgrid line transmission power $\vec{BS}$ contains the battery charging and discharging as well as the capacity, and $\vec{\eta}$ contains the battery charging and discharging efficiency. The constraints can be obtained by describing the maximum value $\vec{X}$ in terms of the respective maximum and minimum values of each element as $\vec{X}_{min} \leq \vec{X} \leq \vec{X}_{max}$. Also note that the system satisfies the following expression:

$$\sum_{i=1}^{N} P_i + P_{IL} = P_L - P_{BS} \quad (28)$$

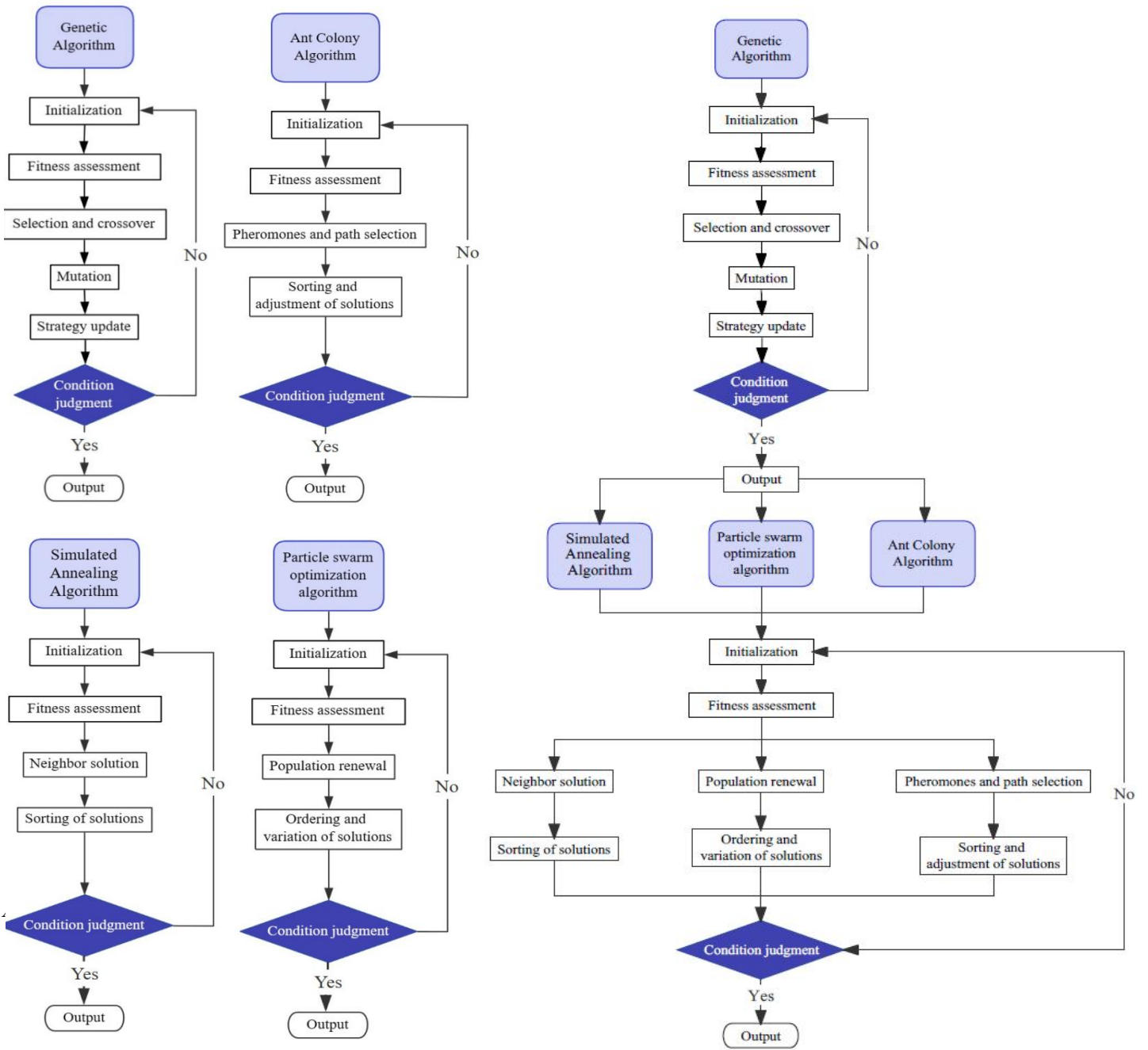

Figure 2. Flowchart of discrete and fusion optimization algorithm

## IV. MULTI-OBJECTIVE ALGORITHM FUSION FOR OPTIMAL SCHEDULING OF MICROGRIDS

In this paper, we use the fusion of multi-objective genetic algorithm with multi-objective particle swarm algorithm, multi-objective ant colony algorithm, and multi-objective simulated annealing algorithm to solve the model, and the general process is shown in Figure 2, and the details of this process as well as the parameters are as follows

### A. Multi-objective genetic algorithm module flow

*1) Initialization:*

①Data Parallel Initialization: Define microgrid system parameters, set model parameters, and constraints. Set genetic algorithm parameters (population size=150, crossover probability=0.95, mutation probability=0.05), and set a small iteration number. Use greedy algorithms or random methods to parallelly initialize the chromosome population. Ensure each chromosome adheres to possible constraints, placing it within a valid search space.

②Parallel Chromosome Simulation: Input each chromosome into the simulation model in parallel. For each chromosome, check for any variables that violate constraints and make appropriate adjustments. Compute operational costs, environmental costs, and potential penalties based on the simulation results. Assign a fitness value to each chromosome based on these metrics.

*2) Fitness Assessment:*

①Normalize Fitness Values: Use non-dominated sorting and normalization to process all chromosome fitness values, ensuring a balanced weight among different objectives.

②Evaluate New Chromosomes: Input newly generated chromosomes into the simulation model for evaluation. Assign a fitness value to each new chromosome based on simulation results.

*3) Selection and Crossover:*

①Roulette Selection: Based on each chromosome's normalized fitness value, calculate its probability of being selected as a parent. Use the roulette wheel method to select high-fitness chromosomes for the crossover and mutation stages.

②Multi-point Crossover Operations: Based on the preset crossover probability, randomly select two chromosomes as parents. Randomly select multiple crossover points on these chromosomes and exchange gene segments between them to produce new offspring chromosomes.

*4) Mutation:*

Decrease mutation rate when population diversity is high and increase it otherwise. For selected chromosomes, randomly choose gene positions and apply bit mutation to produce mutated chromosomes.

*5) Strategy Update:*

① Introducing Elite Strategies:

Select the highest fitness chromosome or a set of chromosomes from the current population. Ensure these elite chromosomes are retained in the next generation, unaffected by crossover and mutation.

② Assessing Population Diversity:

Evaluate the entire population's diversity using gene diversity and fitness diversity. If population diversity drops to a predefined low level, consider reinitializing part of the population to reintroduce diversity.

*6) Stop Condition Judgment:*

Check if the maximum predefined iteration number has been reached. Evaluate the fitness value of the best chromosome in the population, checking for significant improvements in recent iterations.

*7) Output:*

If any termination criteria are met, collate and present the preliminary optimization scheduling results. If not met, revert to step 2 for the next iteration.

*B. Simulated Annealing Algorithm*

*1) Initialization:*

① Data Initialize in Parallel: Define microgrid system parameters, set model parameters and constraints. Initialize simulated annealing parameters: initial temperature, cooling coefficient, termination temperature. Use genetic algorithm results to initialize a starting solution within predefined search space.

② Evaluate the Initial Solution: Introduce the solution to the microgrid simulation model. Compute operational costs, environmental costs, and potential penalties. Assign a fitness value based on these metrics.

*2). Fitness assessment:*

Implement max-min normalization on all fitness values. Ensure balanced weightage among objectives in multi-objective optimization.

*3) Neighbor solution:*

① Determine Neighbor Solution: Generate "neighbor" solutions near the current solution by altering certain variables.

② Evaluate and Update: Input neighbor solutions into the microgrid simulation model. Compute operational costs, environmental costs, and potential penalties for each. Assign fitness values based on these metrics. Update the solution based on simulated annealing criteria (e.g., Metropolis criterion).

*4) Sorting of solutions:*

Rank all solutions using a non-dominated sorting strategy. Store Pareto frontier solutions and remove non-Pareto optimal ones. If archived solutions exceed capacity, retain representative solutions based on crowding distance.

*5) Stop condition judgment:*

Check if temperature is below the termination temperature. Analyze fitness values of best archived solutions for significant recent improvements.

*6) Output:*

If the termination condition is met, sort out and output the final optimal scheduling results, including solutions on the Pareto front, associated operating costs, environmental costs, etc. If the termination condition is not satisfied, return to step 4 and continue to the next iteration.

*C. Particle Swarm Optimization Algorithm*

*1) Initialization:*

①Data Initialize in Parallel: Define microgrid system parameters, set model parameters, and constraints. Configure MOPSO algorithm parameters (particle number=150, iteration number=300, speed limit is -10% to 10% of space range). Use genetic algorithm results to initialize the particle swarm, ensuring each particle's position and velocity lie within the predefined search space and velocity range.

②Parallel Particle Simulation: Simultaneously input multiple particles into the microgrid simulation model. For each particle, check if any of its variables violate any constraints and make appropriate corrections. Compute operational costs, environmental costs, and potential penalties based on the simulation results. Assign fitness values to each particle based on these metrics.

*2) Fitness Assessment:*

Implement max-min normalization on all fitness values. Ensure balanced weightage among objectives in multi-objective optimization.

*3) Population Renewal:*

① Produce Offspring Populations: Based on each particle's

normalized fitness value, use methods like roulette wheel selection to decide which particles will reproduce. Execute crossover and mutation operations for these particles to produce a new generation of particles.

② Determine Individual Extreme Pbest: For each particle, compare its current solution with its historical best (pbest). If the current solution is superior in some objectives, update its pbest value.

*4) Ordering and Variation of Solutions:*

① Hierarchical Sorting and Archiving: Rank all particles using a non-dominated sorting strategy. Store Pareto frontier solutions and remove non-Pareto optimal ones. If archived solutions exceed capacity, retain representative particles based on crowding distance.

② Determine Global Optimal Gbest: From the archived Pareto optimal solutions, use strategies like roulette wheel selection to pick a solution as the global best (gbest).

③ Dynamic Small Probability Variation: Adjust mutation probability or perturbation intensity dynamically based on the current iteration number or other criteria. Randomly select 25% of the particles and slightly perturb their positions to introduce population diversity.

*5) Stop Condition Judgment:*

Check if the maximum predefined iteration number has been reached. Analyze the best solutions' fitness values in the population, checking for significant improvements in recent iterations.

*6) Output:*

If termination criteria are met, collate and present the final optimization scheduling results, including Pareto frontier solutions, associated operational costs, environmental costs, etc. If not met, revert to step 2 for the next iteration.

*D. Ant Colony Algorithm*

*1) Initialization:*

①Data Parallel Initialization: Define microgrid system parameters, set model parameters, and constraints. Configure ant colony algorithm parameters (number of ants=150, iteration number=300, pheromone evaporation coefficient=0.3, pheromone concentration=0.2). Use genetic algorithm results to initialize ant starting positions, ensuring each ant's position is within the predefined search space.

②Parallel Ant Simulation: Simultaneously introduce multiple ants into the microgrid simulation model. For each ant, verify if the chosen path violates any constraints and make appropriate adjustments.

Compute operational costs, environmental costs, and potential penalties for each ant's chosen path. Assign a fitness value to each path based on these metrics.

*2) Fitness Assessment:*

Use max-min normalization to process all fitness values, ensuring a balanced weight among different objectives in multi-objective optimization.

*3) Pheromones and Path Selection:*

①Pheromone Update: Update pheromone concentrations based on the paths chosen by ants and their fitness. Paths with higher fitness get a higher pheromone increment.

②Select New Paths: Ants select new paths based on current pheromone concentrations. The probability of path selection is directly proportional to pheromone concentration and fitness.

③Determine the Global Optimal Solution: From the archived Pareto optimal solutions, select one or multiple solutions as the global best solution.

*4) Sorting and Adjustment of Solutions:*

①Nondominated Sorting and Archiving: Rank all found solutions using nondominated sorting. Store solutions on the Pareto frontier and remove non-Pareto optimal ones. If archived solutions exceed the set capacity, retain representative solutions based on crowding distance.

②Dynamic Adjustment of Pheromone Strength: Adjust the pheromone evaporation coefficient or pheromone strength dynamically based on the current iteration number or other criteria.

*5) Stop Condition Judgment:*

Check if the maximum predefined iteration number has been reached. Analyze the fitness values of the best solutions in the archive, checking for significant improvements in recent iterations.

*6) Output:*

If any termination criteria are met, collate and present the final optimization scheduling results, including solutions on the Pareto frontier, associated operational costs, and environmental costs. If not met, revert to step 2 for the next iteration.

*E. Optimization Algorithm*

*1) Integrated Algorithm Structure Overview:*

①Genetic Algorithm (GA) Entry Point: The process begins with the Genetic Algorithm, which initializes and pre-processes the data. The best solutions or populations derived from GA serve as the starting point for the subsequent algorithms.

②Simulated Annealing (SA): Utilizes the GA output to refine the solution space and find an optimal or near-optimal solution.

③Ant Colony Optimization (ACO): Takes the GA output to guide ants in searching the solution space based on pheromone trails and fitness.

④Particle Swarm Optimization (PSO): Uses the GA output to guide particles in exploring the solution space based on individual and global best solutions.

⑤Result Comparison & Analysis: The solutions from SA, ACO, and PSO are compared and analyzed to derive the most suitable or integrated solution.

*2）Advantages of the Integrated Algorithm:*

①Diversity of Solutions: The combined approach explores a broader solution space, ensuring diverse solutions.

②Robustness: The multi-algorithm approach reduces the risk of settling on local optima.

③Flexibility: Provides multiple solutions, catering to various decision-making criteria.

④Efficiency: GA's entry point offers a good starting solution, speeding up convergence for subsequent algorithms.

⑤Validation: Multiple algorithms arriving at similar solutions serve as a quality check.

⑥Adaptability: The model can be tailored to give more weight to one algorithm based on the problem's nature.

## V. SIMULATION

### A. Simulation

*1) Simulation System*

The simulation in this paper is based on the MATLAB yalmip+cplex environment. Referencing the improved IEEE-RTS 24-node system in References[12-14], we adopt the following microgrid system.

The peak load is 90kW. The uncontrolled clean energy generation modules operate in maximum power point tracking mode, and their 24-hour output is shown in Figure 3:

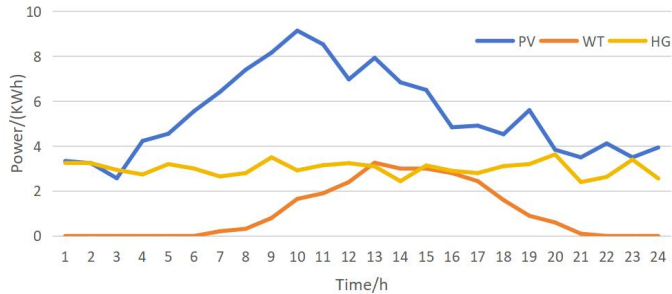

Figure 3: Uncontrolled clean energy generation modules' output over 24 hours

The maximum charge/discharge power of the ES module is set to 5 kW, and the capacity is 25 kW·h. Interruptible loads account for 20% of the total load demand. Other parameters refer to Table 7 and Table 8 below (note the default emission factor of uncontrolled clean energy is 0).

Table 7: Main Parameters of Distributed Power Sources

| Type | Power limit / kW | Management fee / UDS · MW-1 | Service life / y |
|---|---|---|---|
| MT | 45 | 5.6479 | 10 |
| HG | 28 | 3.2143 | 15 |
| DE | 30 | 11.9561 | 10 |
| WT | 15 | 4.1692 | 10 |
| PV | 18 | 1.1973 | 20 |

Table 8. Pollutant Processing Factors and Emission Factors

| Type | Emission Factors(MT) /kg · MW-1 | Emission Factors(DE) /kg · MW-1 | Processing Factors /USD · kg-1 |
|---|---|---|---|
| $CO_2$ | 1.65 | 1.55 | 0.013 |
| $NO_x$ | 0.50 | 19.8 | 3.892 |
| $SO_2$ | 0.01 | 0.51 | 0.892 |

### B. Results and Analysis

*1) Multi-Objective Dispatch Results*

The figure 4 below show the fitted Pareto frontiers of operation and environmental costs obtained by MOGA-MOSA, MOGA-MOPSO, and MOGA-MOACO algorithms, respectively. The results of MOGA-MOPSO and MOGA-MOACO are close, but MOGA-MOSA deviates more, although the trends are consistent among the three. Overall, it is difficult to balance both environmental and operation costs.

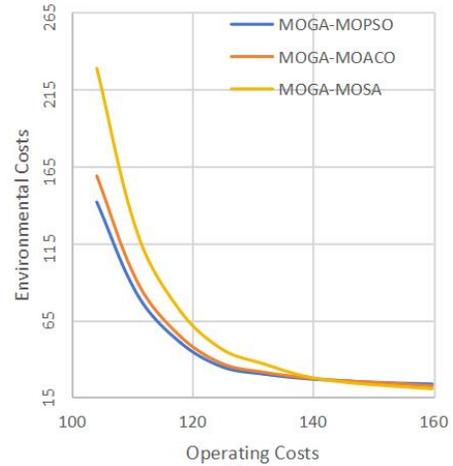

Figure 4: Multi-objective dispatch results

*2) Results of Different Objective Dispatch*

The simulation results of the three algorithms for different objective dispatch are shown in the figure 5 below.

In the figure 5, for example, "Eco Dispatch (MOGA-MOSA) X" indicates the value of X (operation cost) given by the MOGA-MOSA algorithm under environmental dispatch.

The operation and environmental costs obtained under economic and environmental dispatch objectives are given, respectively. The shaded areas map the range of operation and environmental costs under multi-objective dispatch. Although there are some differences in the regions represented by the results of the three algorithms, the overlap is still relatively high. All of them exhibit the regularity that operation costs are lower under economic dispatch while environmental costs are lower under environmental dispatch. In fact, diesel generators have the characteristics of low power generation costs but high pollutant emissions, while micro gas turbines are the opposite. Thus, we can see that under economic dispatch, the microgrid increases the capacity of diesel generator sets; on the other hand, when adopting environmental dispatch, the microgrid increases the capacity of micro gas turbines.

*3) Environmental Dispatch Results*

Environmental dispatch aims to minimize the operation costs of units as the objective function. The following figure 6 show the environmental dispatch results based on MOGA-MOPSO and MOGA-MOACO algorithms. Since the deviation of MOGA-MOSA from the other two is too large, which may be trapped in local optima, MOGA-MOSA is excluded from consideration.

As shown in the figure 6, the microgrid prioritizes power generation from gas turbines at certain time scales, which coincides with the lower pollution characteristics of gas turbines. In addition, interruptible loads will be triggered during the high electricity demand period of 15:00-19:00 to ensure power supply and demand.

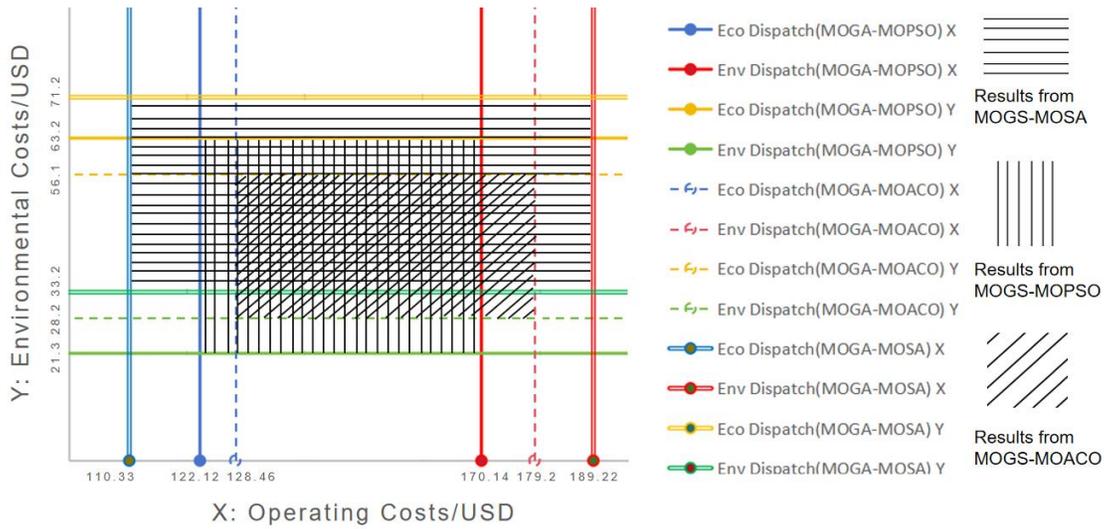
Figure 5: Results of different objective dispatch

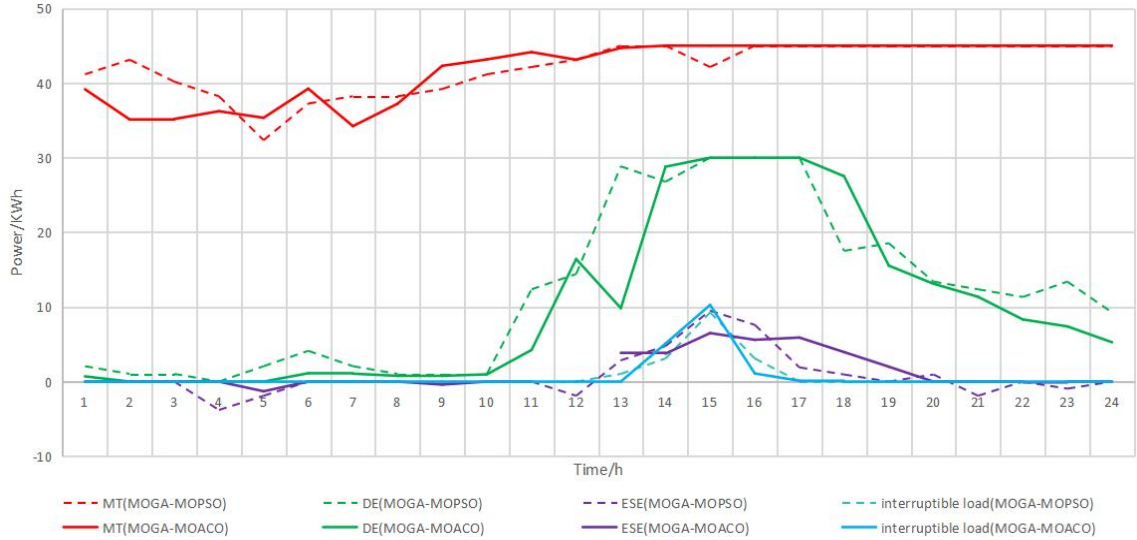
Figure 6: Economic dispatch results

VI. CONCLUSION

With the increasing global attention on renewable and clean energy, research and implementation of microgrids have become especially important. Through the study of multi-objective optimization models for microgrids in this paper, we can draw the following main conclusions:

Tradeoff between power generation efficiency and environmental factors: This paper explores in depth the relationship between operation costs and environmental costs of microgrids through multi-objective optimization models. The results show that pursuing the lowest operation costs may increase environmental costs, and vice versa. Therefore, true optimization should seek a balance point between the two.

Role of different energy sources: Diesel generators and micro gas turbines play different roles in microgrids. Diesel generators have low power generation costs but high pollutant emissions, while micro gas turbines are the opposite. This highlights the need to comprehensively consider various energy sources in microgrid design and operation to achieve both economic and environmental goals.

Importance of multi-objective optimization: Using different optimization algorithms, such as MOGA-MOSA, MOGA-MOPSO and MOGA-MOACO, can produce different dispatch results. These algorithms provide a framework to help decision makers make optimal decisions based on actual situations and policy requirements.

Key role of battery storage modules: To improve microgrid reliability, considering the existence of uncontrolled clean energy, battery storage modules play an indispensable role in the system. They can not only act as an energy buffer to ensure stable power supply, but also charge and discharge on demand to further optimize operation costs.

Future outlook for microgrids: With technological advances and increasing demand for clean energy, operation and management of microgrids will become more intelligent and automated. In the future, multi-objective optimization models will be more complex yet more effective, providing a solid foundation for the sustainable and stable development of microgrids.

Overall, this study provides in-depth insights and practical guidance for microgrid design and operation. It is hoped that the models and technologies can be further improved in the future, making microgrids an important tool for achieving clean, efficient and economical power supply.